\documentclass[letterpaper, 10 pt, conference]{ieeeconf}  

\IEEEoverridecommandlockouts                              

\usepackage{amsmath,graphicx,hyperref}
\usepackage{multirow}
\usepackage{multicol}
\usepackage{arydshln} 
\usepackage{amsfonts}
\usepackage{cite}

\usepackage{todonotes}

\newcommand{\lircsep}{\cdashline{2-12}[0.8pt/2pt]} 
\newcommand{\lircsepab}{\cdashline{1-6}[0.8pt/2pt]}

\title{XD-RCDepth: Lightweight Radar-Camera Depth Estimation with Explainability-Aligned and Distribution-Aware Distillation
}

\author{
Huawei Sun$^{1,2}$, Zixu Wang$^{1,2}$, Xiangyuan Peng$^{1,2}$, Julius Ott$^{1,2}$,\\
Georg Stettinger$^{2}$, Lorenzo Servadei$^{1}$, Robert Wille$^{1}$
\thanks{$^{1}$ Technical University of Munich, Munich, Germany, }
\thanks{$^{2}$ Infineon Technologies AG, Neubiberg, Germany}
}

\begin{document}

\maketitle
\thispagestyle{empty}
\pagestyle{empty}

\begin{abstract}
Depth estimation remains central to autonomous driving, and radar–camera fusion offers robustness in adverse conditions by providing complementary geometric cues. In this paper, we present XD-RCDepth, a lightweight architecture that reduces the parameters by 29.7\% relative to the state-of-the-art lightweight baseline while maintaining comparable accuracy. To preserve performance under compression and enhance interpretability, we introduce two knowledge-distillation strategies: an explainability-aligned distillation that transfers the teacher's saliency structure to the student, and a depth-distribution distillation that recasts depth regression as soft classification over discretized bins. Together, these components reduce the MAE compared with direct training with 7.97\% and deliver competitive accuracy with real-time efficiency on nuScenes and ZJU-4DRadarCam datasets. Code: \url{https://github.com/harborsarah/XD_RCDepth}
\end{abstract}

\section{Introduction}
Despite rapid progress in autonomous driving, depth estimation remains critical and comparatively underexplored. The task typically involves predicting a dense depth map either from an RGB image~\cite{bts,iebins,unidepth,p3depth} or from an RGB image augmented with a sparse LiDAR depth map~\cite{lidar1,lidar2,lidar3}. Purely camera-based approaches are ill-posed since RGB provides no direct geometric measurements, often yielding biased estimates. Incorporating LiDAR points as a geometric prior mitigates this issue and improves accuracy. However, LiDAR is costly and produces high-bandwidth data, creating storage and computation burdens that can impair real-time performance. Additionally, both cameras and LiDAR degrade under adverse weather~\cite{Adverse_survey}. 
For example, cameras perform poorly at night, while LiDAR is strongly affected by fog and snow. 
Motivated by advances in automotive millimeter-wave radar, which is relatively cost-effective and more robust in poor visibility, recent work~\cite{radarnet,getup,RadarCam-Depth,cafnet,tride} has increasingly explored depth estimation from RGB images in combination with sparse radar-derived depth.

Although radar has the advantages noted above, the radar point clouds in the nuScenes dataset are markedly sparser and noisier than LiDAR due to limited elevation resolution, which effectively removes height information from the radar returns \cite{nuscenes}. To compensate, many methods adopt two-stage pipelines that first generate a quasi-dense radar depth map and then refine it with RGB images, or they introduce increasingly complex network designs \cite{rc-pda,radarnet,cafnet,RadarCam-Depth,getup,tride}. Both strategies increase latency and memory usage, undermining real-time deployment. This motivates compact, faster models that retain accuracy.

LiRCDepth \cite{lircdepth} is a pioneering lightweight radar–camera depth framework that improves a compact student by transferring knowledge from a high-capacity teacher via knowledge distillation, a technique originally developed for image classification \cite{kd_ckass1}.
However, its distillation strategy has notable limitations. First, cross-modal feature distillation between image and radar features constrains the student design because it requires channel size alignment at each encoder level; additional affinity modules are therefore needed to project student features to the teacher’s channel dimensions. Second, inter-depth distillation is tightly coupled to the teacher architecture, since many models expose only the final depth prediction without intermediate maps. Third, the proposed distillation signals remain largely superficial and do not engage with model explainability.


Explainable Artificial Intelligence (XAI) has advanced rapidly as a means to elucidate the behavior of black-box neural networks and to clarify why models make particular decisions \cite{xai,xai1}. Several studies distill explainability cues for image classification, improving both the transparency of the student and its accuracy \cite{xai_kd1,xai_kd2}. However, to the best of our knowledge, explainability-aware distillation has not yet been explored for dense prediction tasks such as depth estimation.
\begin{figure*}[ht]
\vspace{0.18cm}
\centering
\includegraphics[width=0.99\textwidth]{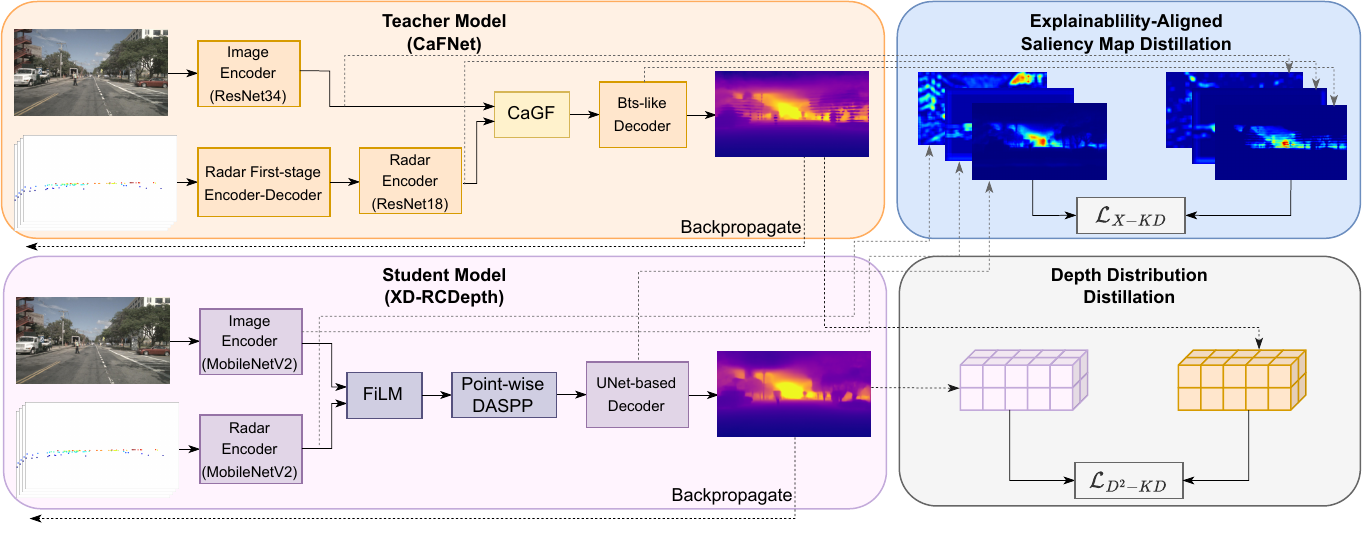}

   \caption{Overview of XD$^{2}$-RCDepth and the proposed distillation objectives. In explainability-aligned distillation (X-KD), Grad-CAM saliency maps from selected student and teacher layers are aligned by minimizing $\mathcal{L}_{\mathrm{X\text{-}KD}}$.  In depth-distribution distillation (D$^{2}$-KD), continuous depth predictions are discretized into per-pixel distributions, and the student is trained to match the teacher by minimizing $\mathcal{L}_{\text{D}^2\text{-KD}}$.
   }
\label{fig:model}
\end{figure*}

To address these limitations, we present XD-RCDepth, a lightweight radar–camera depth estimation architecture that reduces the parameter count by 29.7\% relative to~\cite{lircdepth} while maintaining comparable accuracy. For multimodal fusion, we introduce a compact Feature-wise Linear Modulation (FiLM) scheme that improves performance with negligible parameter overhead. To further preserve accuracy during model compression, we develop two complementary knowledge distillation strategies. First, an explainability-aware distillation aligns the saliency maps of selected intermediate features between the teacher and the student, thereby transferring the teacher’s saliency structure. Second, motivated by the observation that classification is often easier to optimize than regression~\cite{dorn}, we discretize depth into bins and distill the depth distributions by matching the teacher’s and student’s probabilities within each bin. Applied to LiRCDepth, these techniques yield additional performance gains. To the best of our knowledge, this is the first work to leverage explainability for knowledge distillation in dense prediction. 
Our contributions are as follows:
\begin{itemize}
\item A lightweight radar–camera depth estimation framework with an efficient FiLM fusion module is proposed.
\item Two knowledge–distillation methods are proposed to transfer the explainability and the depth distribution from the teacher to the student.
\item Our model is evaluated on nuScenes~\cite{nuscenes} and ZJU-4DRadarCam~\cite{RadarCam-Depth}, achieving comparable results with other heavy-weight architectures.
\end{itemize}

\label{sec:intro}

\section{Related Work}
\subsection{Radar Camera Depth Estimation}
Prior work primarily addresses radar sparsity by employing two-stage networks to predict dense depth~\cite{rc-pda,radarnet,RadarCam-Depth,cafnet}. In these pipelines, the first stage produces a quasi-dense depth map from sparse radar, which is then combined with the RGB image in a second stage to generate the final depth. Although effective, these systems are typically large, and the two-stage design degrades real-time performance. 
Single-stage alternatives such as GET-UP incorporate point-cloud upsampling as an auxiliary task and employ more elaborate radar feature extractors, but at the cost of substantially slower inference \cite{getup}. More recently, language-augmented approaches integrate textual cues via multimodal large language models, which require per-frame description generation and introduce additional latency \cite{tride}. In parallel, LiRCDepth pursues a lightweight radar–camera framework through knowledge distillation from a large teacher to a compact student \cite{lircdepth}. However, its distillation is predominantly feature-level, which enforces channel-dimension alignment across encoder stages and thus limits architectural flexibility or necessitates extra adapters; moreover, it does not explicitly transfer explainability, leaving interpretable guidance underexploited.

\subsection{Knowledge Distillation and Explainable AI} 
Knowledge distillation (KD) transfers knowledge from a high-capacity teacher to a compact student by matching softened teacher outputs under temperature scaling \cite{kd_ckass1}. Subsequent research has enriched supervision beyond logits: FitNets introduce intermediate “hint” layers to guide feature learning \cite{Romero2015FitNets}, while Attention Transfer aligns spatial attention maps to convey where the teacher focuses \cite{Zagoruyko2017AT}. Relation-based KD preserves structural information, including relational distillation across samples \cite{park2019relational} and contrastive representation distillation that aligns instance relations in the embedding space \cite{tian2019contrastive}.
Driven by the need for efficient perception in autonomous driving, KD has been widely adopted beyond classification. For object detection, imitating multi-level pyramid and head features improves both localization and classification \cite{wang2019distilling}. For semantic segmentation, structured distillation aligns pixel-wise and region-level relations, providing stronger guidance than logit matching alone \cite{liu2019structured}. In depth estimation, pairwise or feature-level distillation has been used to transfer supervisory signals from large teachers to lightweight students deployable on mobile hardware \cite{kd_depth1}.

In parallel, Explainable AI (XAI) seeks to elucidate the behavior of black-box neural networks and has advanced rapidly \cite{xai,xai1}. Post-hoc explanations for computer vision models span four families: gradient- or activation-based saliency, perturbation-based analysis, concept-based probing, and attention-based rationales. Class Activation Mapping (CAM) and Gradient-weighted CAM (Grad-CAM) highlight discriminative regions via feature–gradient weighting, with Grad-CAM++ improving localization \cite{Zhou2016CAM,gradcam,Chattopadhyay2018GradCAMpp}. Perturbation and attribution methods offer complementary, often model-agnostic or axiomatic views, exemplified by Local Interpretable Model-agnostic Explanations (LIME), SHapley Additive exPlanations (SHAP), and Integrated Gradients (IG) \cite{Ribeiro2016LIME,Lundberg2017SHAP,Sundararajan2017IG}. Testing with Concept Activation Vectors (TCAV) aligns explanations with human-interpretable concepts \cite{Kim2018TCAV}. Although attention is sometimes presented as a rationale, its faithfulness is debated; attention rollout aggregates signals across layers to improve interpretability \cite{Jain2019AttentionNotExplanation,Abnar2020AttentionRollout}.

Several studies integrate XAI into KD by aligning teacher and student saliency, typically CAM- or gradient-based maps, or by weighting losses with explanation-derived importance in image classification, encouraging students to focus on task-relevant regions \cite{xai_kd1,xai_kd2}. To our knowledge, such XAI-guided KD has not been systematically explored for dense prediction tasks like depth estimation. We adopt this paradigm by enforcing Grad-CAM consistency at selected layers and complement it with a depth-distribution KD objective.

\label{sec:related}

\section{Methodology}
This section first introduces the model architecture. Afterwards, the explainability-aligned saliency map distillation and the depth distribution distillation methods are presented. At last, the loss functions are summarized.  

\subsection{Model Architecture}
As illustrated in Fig.~\ref {fig:model}, to remain comparable with LiRCDepth~\cite{lircdepth}, we adopt CaFNet~\cite {cafnet} as the teacher network. The teacher employs a ResNet-34~\cite{resnet} backbone for the image stream and extracts features at five spatial scales $\{\frac{1}{2^{i}}\}_{i=1}^{5}$. The radar stream is processed in two stages: the first produces a coarse depth map together with a radar confidence map; the second refines radar features by utilizing the results from the first stage. During decoding, radar and image features are fused by Confidence-aware Gated Fusion (CaGF) blocks and then upsampled with a BTS-style decoder~\cite{bts}.

Our student, XD-RCDepth, is built on MobileNetV2~\cite{mobilenetv2} backbones for both modalities and extracts five image features and five radar features at matching scales. At each scale, we propose lightweight FiLM fusion, where per-channel affine parameters are predicted from the radar features and modulate the image features:
\begin{equation}
    \gamma = \text{Conv}_{1\times 1}(f_{r}),\qquad
    \beta = \text{Conv}_{1\times 1}({f_{r}}),
\end{equation}
\begin{equation}
    f_{fuse} = (1+\gamma)\odot f_{i} + \beta,
\end{equation}
Here, $f_{r}$ and $f_{i}$ denote the radar and image features at a given scale, $\gamma,\beta\in \mathbb{R}^{C\times 1\times 1}$ are per-channel scaling and shift coefficients, $\odot$ denotes element-wise multiplication.

To strengthen the decoder with minimal overhead, we adapt Dense Atrous Spatial Pyramid Pooling (DASPP) into a point-wise DASPP module. Following the efficiency rationale of point-wise ($1{\times}1$) convolution \cite{depth_conv}, we replace costly $3{\times}3$ atrous kernels with parallel $1{\times}1$ branches evaluated at stage-specific dilation rates and then aggregate them via concatenation and a final $1{\times}1$ squeeze. This design preserves the multi-scale receptive field of DASPP while substantially reducing parameters and FLOPs. We insert the block at the $\tfrac{1}{32}$, $\tfrac{1}{16}$, and $\tfrac{1}{8}$ decoder stages. Together with FiLM fusion, the point-wise DASPP yields a compact yet expressive student, delivering broader context at negligible computational cost.

Finally, to transfer knowledge from the teacher to the student, we introduce two complementary distillation objectives: Explainability-Aligned Saliency Map Distillation (X-KD), which aligns saliency maps at selected layers, and Depth-Distribution Distillation (D$^2$-KD), which matches the teacher’s soft depth bin distribution. These objectives are detailed in the following section.

\subsection{Explainability-Aligned Saliency Map Distillation}
For a distilled layer $l\in\mathcal{L}$, the student and teacher features are
$F_{l}^{S},F_{l}^{T}\!\in\!\mathbb{R}^{C_{l}\times H_{l}\times W_{l}}$.
Let $\phi^{(\cdot)}$ be a scalar objective (e.g., the mean of the predicted depth) used only to compute gradients. The saliency maps are generated through the Grad-CAM~\cite{gradcam} manner, where weights and maps are
\begin{equation}
\begin{split}
\alpha_{l,c}^{(\cdot)}
&= \frac{1}{H_{l}W_{l}}
  \sum_{i=1}^{H_{l}}\sum_{j=1}^{W_{l}}
  \frac{\partial\,\phi^{(\cdot)}}{\partial\,F_{l,c}^{(\cdot)}(i,j)},\\
\mathrm{Map}_{l}^{(\cdot)}
&= \operatorname{ReLU}\!\Big(
  \sum_{c=1}^{C_{l}} \alpha_{l,c}^{(\cdot)}\, F_{l,c}^{(\cdot)}
\Big)
\in \mathbb{R}^{H_{l}\times W_{l}},
\end{split}
\end{equation}
where $(\cdot)\in\{S,T\}$ indexes the student or teacher.
We then flatten (denoted as $\operatorname{vec}(\cdot)$)  and $\ell_{2}$-normalize of each map:
\begin{equation}
\tilde{\mathbf a}_{l}^{(\cdot)}
= \frac{\operatorname{vec}\!\big(\mathrm{Map}_{l}^{(\cdot)}\big)}
       {\big\|\operatorname{vec}\!\big(\mathrm{Map}_{l}^{(\cdot)}\big)\big\|_{2}+\varepsilon}
\in \mathbb{R}^{H_{l} W_{l}}.
\end{equation}

The per-layer cosine loss and the final explainability-aligned distillation loss are
\begin{equation}
\ell_{\mathrm{map},l}
= 1 - \big\langle \tilde{\mathbf a}_{l}^{S}, \tilde{\mathbf a}_{l}^{T} \big\rangle,
\qquad
\mathcal{L}_{\mathrm{X\text{-}KD}}
= \frac{1}{|\mathcal{L}|}\sum_{l\in\mathcal{L}} \ell_{\mathrm{map},l}.
\end{equation}

In practice, teacher maps are detached so that gradients from $\mathcal{L}_{\mathrm{X\text{-}KD}}$ are backpropagated only to the student.
\subsection{Depth Distribution Distillation}
We use the teacher’s predicted depth as soft labels to guide the student, following the idea of distilling a teacher’s soft targets in classification \cite{kd_ckass1}. We first partition the continuous depth range $[d_{\min}, d_{\max}]$ into $B$ bins with centers $\{c_i\}_{i=1}^{B}$, where $i$ indexes the $i$-th depth bin. 
For pixel $p$, let the teacher and student depths be $d^{T}(p)$ and $d^{S}(p)$, and let the temperature be $\tau>0$.

We compute absolute deviations to each bin center and form per-bin logits:
\begin{equation}
\Delta_{i}^{(\cdot)}(p)=\bigl|\,d^{(\cdot)}(p)-c_i\,\bigr|,
\qquad
z_{i}^{(\cdot)}(p)= -\,\Delta_{i}^{(\cdot)}(p),
\end{equation}
where $(\cdot)\in\{S,T\}$ denotes the student or teacher.

A temperature-scaled softmax then produces per-pixel categorical distributions over bins:
\begin{equation}
\resizebox{0.9\columnwidth}{!}{$
\mathbf p^{S}(p)=\mathrm{softmax}\!\big(\mathbf z^{S}(p)/\tau\big),\quad
\mathbf q^{T}(p)=\mathrm{softmax}\!\big(\mathbf z^{T}(p)/\tau\big).
$}
\end{equation}

Finally, we minimize the forward KL divergence from the teacher to the student, averaged over valid pixels $\Omega$, with the usual $\tau^{2}$ prefactor:
\begin{equation}
\mathcal{L}_{\text{D}^2\text{-KD}}
= \frac{\tau^{2}}{|\Omega|}
\sum_{p\in\Omega}\;\sum_{i=1}^{B}
q^{T}_{i}(p)\,\log\!\frac{q^{T}_{i}(p)}{p^{S}_{i}(p)}\,.
\end{equation}

\subsection{Loss Functions}
To supervise depth estimation, we first accumulate LiDAR points from neighboring frames to form a dense depth map $\mathbf{D}_{\mathrm d}$. In addition, a single–scan LiDAR depth map $\mathbf{D}_{\mathrm s}$ is used during training, following~\cite{lircdepth}. The depth loss is defined as the mean absolute error (MAE) over the valid pixels of each supervision source:
\begin{equation}
\resizebox{0.9\columnwidth}{!}{$
\mathcal{L}_{\text{Depth}}
= \dfrac{1}{|\Omega_{\mathrm s}|}\;\sum\limits_{x\in\Omega_{\mathrm s}}
  \bigl|\mathbf{D}_{\mathrm s}(x)-\hat{\mathbf{D}}(x)\bigr|
+ \dfrac{1}{|\Omega_{\mathrm d}|}\;\sum\limits_{x\in\Omega_{\mathrm d}}
  \bigl|\mathbf{D}_{\mathrm d}(x)-\hat{\mathbf{D}}(x)\bigr| ,
$}
\end{equation}
where $\hat{\mathbf{D}}$ denotes the predicted depth, and
$\Omega_{\mathrm s}$, $\Omega_{\mathrm d}$ are the sets of pixels where
$\mathbf{D}_{\mathrm s}$ and $\mathbf{D}_{\mathrm d}$ are valid, respectively.

The overall training objective combines the supervised depth loss and the two distillation terms:
\begin{equation}
\mathcal{L}
= \lambda_{1}\,\mathcal{L}_{\text{Depth}}
+ \lambda_{2}\,\mathcal{L}_{\mathrm{X\text{-}KD}}
+ \lambda_{3}\,\mathcal{L}_{\text{D}^{2}\text{-KD}},
\end{equation}
where $\lambda_{1},\lambda_{2},\lambda_{3}\!\ge\!0$ are weighting factors that balance the contributions of the respective loss terms.

\label{sec:methods}

\section{Experiments}
\subsection{Datasets and Implementation Details}
We experiment on nuScenes~\cite{nuscenes} and ZJU-4DRadarCam~\cite{RadarCam-Depth} datasets to validate our approach under both 3D- and 4D-radar regimes.
For nuScenes, we construct the supervision depth $\mathbf{D}{\mathrm{s}}$ using the same accumulation procedure as in~\cite{cafnet}. For ZJU\textendash 4DRadarCam, we directly use the provided dense depth map as $\mathbf{D}{\mathrm{s}}$. All models are trained on a single NVIDIA L40 GPU with a batch size of $8$.

For computing $\mathcal{L}_{\mathrm{X\text{-}KD}}$, we distill from three layers (image encoder, radar encoder, decoder) at the $\tfrac{1}{16}$ scale. Additionally, the loss weights are set to $\lambda_1=1.0$, $\lambda_2=0.5$, and $\lambda_3=0.5$.

\subsection{Evaluation Metrics}
Table \ref{table:metrics} summarizes the evaluation metrics. Let $\hat{\mathbf{D}}$ and $\mathbf{D}_{gt}$ denote the predicted and ground-truth depth maps, respectively, and let $\Omega$ be the set of pixels where $\mathbf{D}_{gt}$ is valid.

\begin{table}[ht]
\vspace{-2mm}
\caption{Metrics definition for depth estimation task.}
\vspace{-1mm}
\centering
\begin{tabular}{l||cccc}
\hline
 & Definition\\ \hline
MAE  &  $\frac{1}{|\Omega|}\sum_{x\in \Omega} |\hat{\mathbf{D}}(x)-\mathbf{D}_{gt}(x)|$      \\ 
RMSE    &   $ (\frac{1}{|\Omega|}\sum_{x\in \Omega} |\hat{\mathbf{D}}(x)-\mathbf{D}_{gt}(x)|^{2})^{1/2} $  \\
AbsRel       & $\frac{1}{|\Omega|}\sum_{x\in \Omega} |\hat{\mathbf{D}}(x)-\mathbf{D}_{gt}(x)|/\mathbf{D}_{gt}(x)$ \\
log10  & $\frac{1}{|\Omega|}\sum_{x\in \Omega} |\log_{10}\hat{\mathbf{D}}(x)-\log_{10}\mathbf{D}_{gt}(x)|$ \\
RMSElog & $\sqrt{\frac{1}{|\Omega|}\sum_{x\in \Omega} ||\log_{10}\hat{\mathbf{D}}(x)-\log_{10}\mathbf{D}_{gt}(x)||^{2}}$\\
$\delta_{n}$ threshold   &   $\delta_{n}=|\{\hat{\mathbf{D}}(x): max(\frac{\hat{\mathbf{D}}(x)}{\mathbf{D}_{gt}(x)}, \frac{\mathbf{D}_{gt}(x)}{\hat{\mathbf{D}}(x)})< 1.25^{n}\}|/|\Omega|$        \\
 \hline
\end{tabular}
\label{table:metrics}
\vspace{-3.5mm}
\end{table}

\subsection{Quantitative Results}
\begin{table*}[ht]
\vspace{0.18cm}

\centering
\caption{Performance Comparison on nuScenes Official Test Set. }
\label{tab:exp}
\resizebox{17.5cm}{!} 
{
\centering

\begin{tabular}{c||c|c|c|cccccccc}
\hline
\multirow{2}{*}{Eval Distance} & \multirow{2}{*}{Method} & \multicolumn{2}{c|}{Model} & \multicolumn{8}{c}{Metrics} \\ \cline{3-12} 
                                    &                         & Params$\downarrow$      & Runtime$\downarrow$   & MAE $\downarrow$ & RMSE $\downarrow$ & AbsRel $\downarrow$  &log10 $\downarrow$ & RMSElog $\downarrow$ & $\delta_{1}$ $\uparrow$& $\delta_{2}$ $\uparrow$ & $\delta_{3}$ $\uparrow$  \\ \hline
\multirow{7}{*}{50m}                & RadarNet   \cite{radarnet}  & 8.39M+14.41M  & 0.336s+0.042s & 1.706 & 3.742 & 0.103 & 0.041 & 0.170 & 0.903 & 0.965 &0.983 \\ 
                                    & CaFNet (T)  & 62.25M & 0.132s & 1.359 & 3.131 & 0.079 & 0.031 & 0.145 & 0.931 & 0.974 & 0.987 \\
                                    \lircsep
                                    & LiRCDepth (w/o KD) & 12.65M & 0.069s & 1.638 & 3.499 & 0.101 & 0.039 & 0.163 & 0.905 & 0.965 & 0.985 \\
                                    & LiRCDepth (KD)      & 12.65M & 0.069s  & 1.514 & 3.330 & 0.092 & 0.036 & 0.155 & 0.916  & 0.969 & 0.986\\
                                    & LiRCDepth (XD$^2$-KD)& 12.65M & 0.069s & \underline{1.479} & \underline{3.256} & \underline{0.084} & \underline{0.035} & \underline{0.150} & \underline{0.919}&\underline{0.971} & \underline{0.987} \\
                                    \lircsep
                                    & XD-RCDepth (w/o KD)& 8.89M & 0.015s & 1.746 & 3.654 & 0.110 & 0.042 & 0.172 & 0.898 & 0.962 & 0.983 \\
                                    & XD-RCDepth (XD$^2$-KD) & 8.89M & 0.015s & \textbf{1.608}  &\textbf{3.511}  &\textbf{0.098}  &\textbf{0.038}  &\textbf{0.161}  &\textbf{0.911}  &\textbf{0.966}  &\textbf{0.984}  \\
                                    \hline
\multirow{7}{*}{70m}                & RadarNet    \cite{radarnet}  & 8.39M+14.41M   & 0.336s+0.042s & 2.073 & 4.591 & 0.105  & 0.043 & 0.181 & 0.896 & 0.962 & 0.981      \\ 
                                    & CaFNet (T) & 62.25M  & 0.132s &  1.673 & 3.928 & 0.082 & 0.033 & 0.154 & 0.923 & 0.971  & 0.986 \\
                                    \lircsep
                                    & LiRCDepth (w/o KD)  & 12.65M  & 0.069s & 2.037 & 4.479 & 0.104 & 0.042 & 0.174 & 0.894 & 0.960 & 0.982 \\
                                    & LiRCDepth (KD)      & 12.65M  & 0.069s & 1.898 & \underline{4.300} & 0.095 & \underline{0.036} & \underline{0.155} & \underline{0.916}  & \underline{0.969} & 0.986\\
                                    & LiRCDepth (XD$^2$-KD)& 12.65M & 0.069s & \underline{1.887} & 4.324 & \underline{0.089} & 0.038 & 0.165 & 0.915 & \underline{0.969}  &\underline{0.987} \\
                                    \lircsep
                                    & XD-RCDepth (w/o KD)& 8.89M & 0.015s & 2.123 &4.591  & 0.114 & 0.044 & 0.182 & 0.889 &0.957  &0.980 \\
                                    & XD-RCDepth (XD$^2$-KD) & 8.89M & 0.015s & \textbf{1.954} &\textbf{4.393}  &\textbf{0.101}  &\textbf{0.040}  &\textbf{0.171}  &\textbf{0.903}  &\textbf{0.962}  &\textbf{0.982}  \\
                                    \hline
\multirow{7}{*}{80m}                & RadarNet  \cite{radarnet}  &8.39M+14.41M  & 0.336s+0.042s & 2.179 & 4.899 & 0.106 & 0.044 & 0.184 & 0.894 & 0.959 & 0.980 \\ 
                                    & CaFNet (T) &62.25M  & 0.132s & 1.763 & 4.184 & 0.083 & 0.034 & 0.156 & 0.921 & 0.970 & 0.985\\
                                    \lircsep
                                    & LiRCDepth (w/o KD) & 12.65M & 0.069s  & 2.152 & 4.801 & 0.105 & 0.043 & 0.177 & 0.892 & 0.959 & 0.981 \\
                                    & LiRCDepth (KD)      & 12.65M  & 0.069s  & 2.009 & \underline{4.617} &0.096 & \underline{0.039} & 0.170 & 0.903  & 0.963 & \underline{0.983}\\
                                    & LiRCDepth (XD$^2$-KD)& 12.65M & 0.069s & \underline{2.005} & 4.676 & \underline{0.090} & \underline{0.039} & \underline{0.169} & \underline{0.906} &\underline{0.964}  &\underline{0.983} \\
                                    \lircsep
                                    & XD-RCDepth (w/o KD)& 8.89M & 0.015s & 2.232 & 4.897 & 0.114 & 0.045 & 0.185 & 0.887 & 0.956 & 0.980\\
                                    & XD-RCDepth (XD$^2$-KD) & 8.89M & 0.015s &\textbf{2.054} &\textbf{4.676}  &\textbf{0.102}  &\textbf{0.041}  &\textbf{0.174}  &\textbf{0.901}  &\textbf{0.961}  &\textbf{0.982}\\
                                    \hline
\end{tabular}
}
\\[1pt]
\scriptsize{T: Teacher model, w/o KD: Student model trained without distillation, KD: distillation methods from~\cite{lircdepth}, XD$^2$-KD: distillation approaches proposed in this paper.}
\end{table*}

\begin{table}[ht]
\centering
\caption{Performance on ZJU-4DRadarCam Test Set. }
\label{tab:exp_zju}
\resizebox{\columnwidth}{!} 
{
\centering

\begin{tabular}{c||c|cccc}
\hline
Eval Distance & Method  & MAE $\downarrow$ & RMSE $\downarrow$ & AbsRel $\downarrow$ & $\delta_{1}$ $\uparrow$ \\ \hline
\multirow{4}{*}{50m}                & RadarNet   \cite{radarnet}  & 1.785 & 3.704 & - & -  \\ 
                                    & CaFNet (T) & 1.044 & 2.722 & 0.083 & 0.901 \\
                                    & XD-RCDepth (w/o KD)& 1.218 & 2.912 & 0.098 & 0.891\\
                                    & XD-RCDepth (XD$^2$-KD)& \textbf{1.155} & \textbf{2.863}  & \textbf{0.094}  & \textbf{0.891} \\
                                    \hline
\multirow{4}{*}{70m}                & RadarNet    \cite{radarnet}  & 1.932 & 4.137 & - & -    \\ 
                                    & CaFNet (T) & 1.125 & 3.078 & 0.084 & 0.900 \\
                                    & XD-RCDepth (w/o KD) & 1.321 & 3.291 & 0.099 &0.889 \\
                                    & XD-RCDepth (XD$^2$-KD)& \textbf{1.248}  & \textbf{3.223}  & \textbf{0.095}  & \textbf{0.890} \\
                                    \hline
\multirow{4}{*}{80m}                & RadarNet  \cite{radarnet}  & 1.979 & 4.309 & - & -\\ 
                                    & CaFNet (T) & 1.149 & 3.195 & 0.084 & 0.899\\
                                    & XD-RCDepth (w/o KD) & 1.351 & 3.414 & 0.099 & 0.889\\
                                    & XD-RCDepth (XD$^2$-KD) & \textbf{1.275} & \textbf{3.341}  & \textbf{0.095}  & \textbf{0.889}  \\
                                    \hline
\end{tabular}
}
\end{table}
We evaluate on nuScenes~\cite{nuscenes} and ZJU\textendash4DRadarCam~\cite{RadarCam-Depth} using the metrics presented in Table~\ref{table:metrics}.  
First, we present the results, evaluated on the nuScenes dataset, in Table~\ref{tab:exp}.
To show the effect of our distillation, we first plug X\text{-}KD and D$^{2}$\text{-}KD into the LiRCDepth teacher–student framework. As shown in Table~\ref{tab:exp}, this yields consistent gains, with the largest improvement at 50\,m, since we use the image-level mean depth as the gradient-based saliency map target, which provides a stable scalar for backpropagation and produces smoother, more informative saliency.
Our XD-RCDepth contains only 8.89\,M parameters, 29.7\% fewer than LiRCDepth~\cite{lircdepth}. Additionally, our model achieves about 66 frames per second, supporting real-time deployment. With X\text{-}KD and D$^{2}$\text{-}KD enabled, XD-RCDepth further reduces MAE relative to its non-distilled counterpart by 7.91\%, 7.96\%, and 7.97\% at 50, 70, and 80\,m.

To further assess our approach, we train and evaluate on ZJU–4DRadarCam, a recent benchmark tailored to radar–camera depth estimation that uses modern 4D imaging radar, yielding denser point clouds with elevation cues than the 3D radar in nuScenes. Because prior work rarely reports on this dataset and public teacher checkpoints are unavailable, we reproduce CaFNet on the ZJU-4DRadarCam dataset (Table \ref{tab:exp_zju}). With X-KD and D$^{2}$-KD enabled, XD-RCDepth achieves relative reductions of 5.17\%, 5.53\%, and 5.63\% in MAE and 1.68\%, 2.07\%, and 2.14\% in RMSE at 50, 70, and 80 meters, respectively, compared with the non-distilled student, confirming the effectiveness of our distillation under 4D radar.

\begin{figure}[hbpt]
    \centering
    \includegraphics[width=0.99\linewidth]{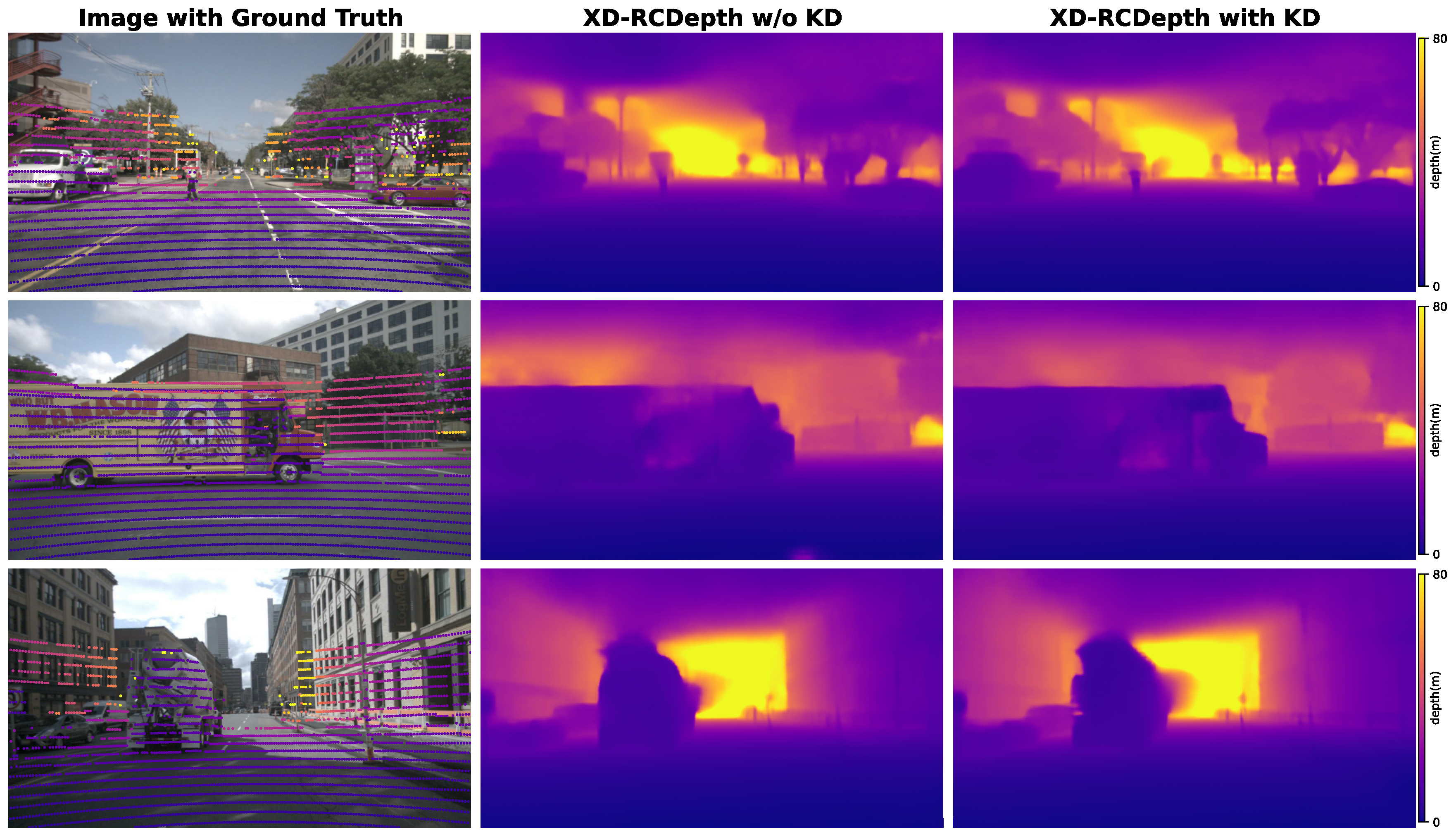}
    
       \caption{Qualitative comparison on the nuScenes dataset between the XD-RCDepth trained with and without distillation.}
    \label{fig:qualitative}
\end{figure}
\subsection{Qualitative Results}
\begin{figure}[hbpt]
    \centering
    \includegraphics[width=0.99\linewidth]{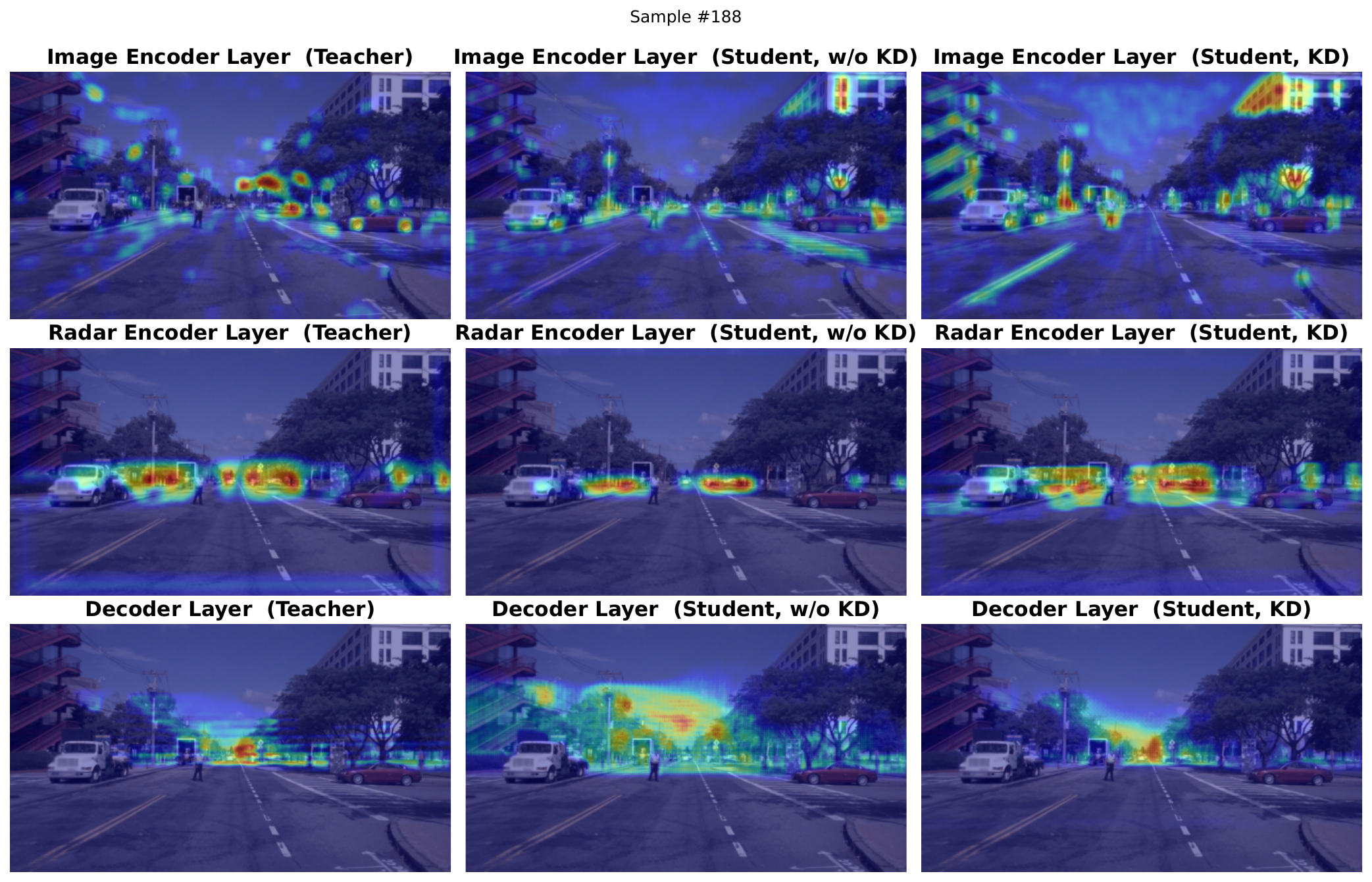}
    
       \caption{Saliency map comparison between the teacher and the student model trained with and without our distillation.}
    \label{fig:cam}
    \vspace{-4mm}
\end{figure}

We first present qualitative results on the nuScenes dataset. As shown in Fig. \ref{fig:qualitative}, the distilled model yields sharper object boundaries and cleaner depth discontinuities than its non-distilled counterpart. We then compare saliency map visualizations from three selected layers for the teacher and for XD-RCDepth trained with and without our distillation (Fig. \ref{fig:cam}). 
Training with the explainability-aligned loss $\mathcal{L}_{\mathrm{X\text{-}KD}}$ yields saliency maps that are sharper and more focused on depth-relevant structures, closer matching the teacher’s attention than the non-distilled counterpart.

\subsection{Ablation Studies}
We perform a series of ablations to quantify the contribution of each proposed component. The results are reported on the nuScenes test set at the 80 m range. We first compare fusion strategies by replacing FiLM with alternative modules and reporting both accuracy and parameter count. We then assess the impact of the Pointwise DASPP by removing it from the decoder. Increasing 0.4M parameters, resulting in an improvement of the MAE by 2.3\%. These models are trained from scratch without any distillation; the results are summarized in Table \ref{tab:ablation_fusion}. 

Finally, we evaluate the distillation mechanisms in isolation by enabling either X-KD or D\textsuperscript{2}-KD individually and reporting their incremental gains in Table \ref{table:ablation_kd}, combining $\mathcal{L}_{\mathrm{X\text{-}KD}}$ and $\mathcal{L}_{\text{D}^2\text{-KD}}$ significantly improves the model accuracy.

\begin{table}[ht]
\centering
\caption{Study of fusion methods and point-wise DASPP.}
\label{tab:ablation_fusion}
\resizebox{0.95\columnwidth}{!} 
{
\centering

\begin{tabular}{c||c|cccc}
\hline
Fusion Method & Params $\downarrow$  & MAE $\downarrow$ & RMSE $\downarrow$ & AbsRel $\downarrow$ & $\delta_{1}$ $\uparrow$ \\ \hline
add & 8.74M & 2.248 & 4.903 & 0.115 & 0.886 \\
concatenate & 10.94M & 2.208 & 4.802 & 0.114 & 0.888 \\
attention & 9.48M & 2.266 & 4.901 & 0.115 & 0.885 \\
FiLM & 8.89M & \textbf{2.232} & \textbf{4.897} & \textbf{0.114} & \textbf{0.887} \\
\lircsepab
FiLM w/o DASPP & 8.45M & 2.285 & 4.913 & 0.117 & 0.872  \\

\hline
\end{tabular}
}
\vspace{-5mm}
\end{table}

\begin{table}[ht]
\centering

\caption{Ablation study on the distillation losses.}
\resizebox{0.95\columnwidth}{!} 
{
\centering
\begin{tabular}{cc||cccc}
\hline
$\mathcal{L}_{\mathrm{X\text{-}KD}}$ & $\mathcal{L}_{\text{D}^2\text{-KD}}$ & MAE $\downarrow$ & RMSE $\downarrow$ & AbsREL $\downarrow$ & $\delta_{1}$ $\uparrow$\\ \hline
- & -  & 2.232 & 4.897 & 0.114 & 0.887  \\
\hline
\checkmark & & 2.114 & 4.756  & 0.108 & 0.892 \\
 &  \checkmark & 2.132 & 4.781 & 0.107  & 0.891\\
\checkmark & \checkmark & \textbf{2.054} & \textbf{4.676} & \textbf{0.102} & \textbf{0.901}  \\

 \hline
\end{tabular}
}
\label{table:ablation_kd}
\vspace{-5mm}
\end{table}
\label{sec:exps}

\section{Conclusion}
We present XD-RCDepth, a lightweight radar–camera depth estimation framework that combines an efficient FiLM fusion module with a point-wise DASPP decoder to improve accuracy at low cost. Compared with the leading lightweight baseline, our model reduces parameters by 29.7\% while delivering comparable performance. To train effectively and enhance interpretability, we introduce two complementary distillation strategies: an explainability-aligned saliency map distillation that transfers the teacher’s saliency patterns to the student, and a depth-distribution distillation that treats the teacher’s predictions as soft labels to guide the student toward calibrated depth estimates. Extensive experiments on the nuScenes and ZJU-4DRadarCam datasets validate the effectiveness of these components. In future work, we will study how the choice of Grad-CAM target and alternative attribution objectives influences the quality of distilled explainability and downstream performance.
\label{sec:concl}

\bibliographystyle{IEEEtran}
\bibliography{main}

\end{document}